\documentclass[letterpaper]{article} 
\usepackage[draft]{aaai25}  
\usepackage{times}  
\usepackage{helvet}  
\usepackage{courier}  
\usepackage[hyphens]{url}  
\usepackage{graphicx} 
\usepackage{natbib}  
\usepackage{caption} 
\usepackage{algorithm}
\usepackage{amsmath}
\usepackage{float}
\usepackage{caption}
\usepackage{subfig}
\usepackage{color}
\usepackage{multirow}
\usepackage{cite}
\usepackage{booktabs}
\usepackage{array}
\usepackage{xcolor}
\usepackage{algpseudocode}
\usepackage{multirow,bm}
\usepackage{cleveref}
\usepackage{amssymb}
\usepackage{mathtools}
\usepackage[none]{hyphenat}
\usepackage{enumitem}

\usepackage{newfloat}
\usepackage{listings}
\DeclareCaptionStyle{ruled}{labelfont=normalfont,labelsep=colon,strut=off} 
\floatstyle{ruled}
\newfloat{listing}{tb}{lst}{}
\floatname{listing}{Listing}
\title{Multiscale fusion enhanced spiking neural network \\ for invasive BCI neural signal decoding}
\author{
    Yu Song\textsuperscript{\rm 1},
    Liyuan Han\textsuperscript{\rm 2},
    Bo Xu\textsuperscript{\rm 1,2,3}\thanks{Corresponding author: xubo@ia.ac.cn},
    Tielin Zhang\textsuperscript{\rm 1,2,3}\thanks{Corresponding author: zhangtielin@ion.ac.cn}
}
\affiliations{
    \textsuperscript{\rm 1}School of Artificial Intelligence, University of Chinese Academy of Sciences \\
    \textsuperscript{\rm 2}Center for Excellence in Brain Science and Intelligence Technology, Chinese Academy of Sciences \\
    \textsuperscript{\rm 3}Institute of Automation, Chinese Academy of Sciences \\
    \textsuperscript{\rm 4}Center for Excellence in Brain Science and Intelligence Technology, Chinese Academy of Sciences
}

\usepackage{bibentry}

\begin{document}

\maketitle

\begin{abstract}
\begin{quote}
Brain-computer interfaces (BCIs) are an advanced fusion of neuroscience and artificial intelligence, requiring stable and long-term decoding of neural signals. Spiking Neural Networks (SNNs), with their neuronal dynamics and spike-based signal processing, are inherently well-suited for this task. This paper presents a novel approach utilizing a Multiscale Fusion enhanced Spiking Neural Network (MFSNN). The MFSNN emulates the parallel processing and multiscale feature fusion seen in human visual perception to enable real-time, efficient, and energy-conserving neural signal decoding. Initially, the MFSNN employs temporal convolutional networks and channel attention mechanisms to extract spatiotemporal features from raw data. It then enhances decoding performance by integrating these features through skip connections. Additionally, the MFSNN improves generalizability and robustness in cross-day signal decoding through mini-batch supervised generalization learning. In two benchmark invasive BCI paradigms, including the single-hand grasp-and-touch and center-and-out reach tasks, the MFSNN surpasses traditional artificial neural network methods, such as MLP and GRU, in both accuracy and computational efficiency. Moreover, the MFSNN’s multiscale feature fusion framework is well-suited for the implementation on neuromorphic chips, offering an energy-efficient solution for online decoding of invasive BCI signals.
\end{quote}
\end{abstract}

\section{Introduction}

Simulating the human brain remains a key objective in neuroscience and artificial intelligence. While Large Language Models (LLMs), such as GPT \cite{achiam2023gpt}, aim to replicate the brain's broad functionality on a general-purpose scale, and brain-inspired neural networks \cite{schmidgall2024brain,zhang2023brain,zhao2023ode} focus on capturing its dynamic complexity, there is also a crucial need to understand and model the brain's inner workings on a microscopic level. Recent advancements in invasive Brain-Computer Interface (BCI) technology allow for the direct recording of spike signals at this microscopic scale. Deep learning models can map these microscopic spike signals to macroscopic behavioral outputs through neural signal decoding. For example, SGLNet \cite{gong2023spiking} converts EEG signals into spike trains, utilizing Spiking Neural Networks (SNNs) to extract topological information and spike-based LSTM units to decode temporal dependencies. Similarly, hand gesture decoding has been achieved by decomposing high-density electromyography signals into motor unit spike trains \cite{chen2020hand}, which are classified to estimate each gesture.

A major challenge in long-term invasive BCI recordings is the issue of data distribution shifts due to factors such as electrode drift and inflammation, which can degrade model generalization. Addressing stable cross-day decoding is thus a critical goal in BCI research. For instance, a DRNN \cite{ran2019decoding} demonstrated high accuracy and robustness in decoding arm velocity during a macaque monkey's reaching task. Attention-based models, such as the Temporal Attention-aware Timestep Selection (TTS) \cite{yang2021selection}, have improved RNN-based neural decoders by selecting key timesteps to enhance accuracy and efficiency. To overcome data limitations, the spatiotemporal Transformer model NDT2 \cite{ye2024neural} leveraged pre-training across sessions, subjects, and tasks, using cross-attention mechanisms and the PerceiverIO architecture to adapt quickly to new sessions with mini-batch labeled data, effectively analyzing diverse neural recordings.

Despite achieving high decoding accuracy, traditional ANN models often suffer from high energy consumption. In contrast, the human brain operates with remarkable energy efficiency. This paper is driven by the need to explore brain-inspired mechanisms for more efficient information processing. As depicted in Fig.\ref{fig1} A, the human brain employs parallel processing pathways, specifically the dorsal and ventral streams, to handle visual inputs \cite{kandel2000principles}. These pathways process signals hierarchically in lower-level brain regions, extracting and integrating multiscale features. The functional differentiation between the pathways allows simultaneous extraction of diverse features, which are integrated in higher-level brain regions to form a unified visual perception. This approach contributes to the brain's exceptional signal processing efficiency \cite{roy2019towards}. Furthermore, the brain transmits neural signals through discrete action potentials, or ``spikes'', leading to low energy consumption. SNNs mimic this spike-based communication, where synaptic connections are activated and adjusted only when spikes occur, offering high biological realism. Consequently, SNNs are more energy-efficient in decoding neural signals compared to ANNs \cite{zhang2023brain}. In the BCI domain, SNNs play a crucial role in reducing energy consumption, enabling high-throughput invasive BCI systems to achieve high performance while being more compact and extending battery life. Such advancements are essential for the clinical application and commercialization of implantable or portable BCI devices \cite{makarov2022toward}.

Inspired by the brain's parallel processing architecture and multiscale feature fusion mechanisms, this paper proposes a method for energy-efficient, invasive cross-day decoding. The main contributions of our work are summarized as follows:

\begin{enumerate}[label={\arabic*)}]
\item We introduce a Multiscale Fusion enhanced Spiking Neural Network (MFSNN), which emulates the brain's visual pathways to achieve efficient signal fusion and feature extraction, thereby improving cross-day decoding performance.
\item Our SNN-based model decodes high-throughput invasive brain signals with reduced energy consumption, offering a practical solution for invasive BCI systems.
\item In two invasive BCI paradigms, i.e., single-hand grasp-and-touch and center-and-out reach tasks, the MFSNN demonstrates the feasibility and robustness of cross-day decoding in BCI systems through mini-batch supervised generalization learning.
\end{enumerate}

\section{Related Works}  
\subsection{Brain-inspired Computing}  
The human brain is the only biological system that demonstrates advanced general intelligence with ultra-low power consumption. Insights from the brain have the potential to propel a narrow AI towards a more general one. Brain-inspired computing (BIC) embraces this concept, introducing a new paradigm of computation and learning, inspired by the fundamental structures and information processing mechanisms of the human brain \cite{liu2024advancing}. It has been found that the parallel processing and multiscale feature fusion are prevalent in brain information processing \cite{zeki2016multiple}, and various parallel subsystems extract different aspects of signal features, with feature fusion occurring both within and between subsystems. For example, the human visual system employs complex and organized strategies of parallel processing, hierarchical fusion, and modularity to transform and interpret visual information \cite{nassi2009parallel}. In the olfactory system, olfactory perception is processed through two parallel pathways within the olfactory bulb\cite{vaaga2016parallel}. Similarly, in the pain system, two parallel pathways, medial and lateral, are responsible for transmitting and processing nociceptive and emotional information, respectively \cite{nociceptive}. Beyond integrating different signal features, the parallel structures in the brain also help reduce operational energy consumption.

The rapid advancements in BCI decoding involve processing multi-channel, high-sampling signals, which can be inefficient with conventional single-network architectures. In contrast, parallel network structures, inspired by brain mechanisms, can handle these signals more efficiently and extract features across various dimensions. However, whether in human visual perception \cite{katsuki2014bottom} or in artificial intelligence's parallel networks, processing high-throughput multi-channel signals requires channel attention mechanisms \cite{zhao2023local} to discern the importance of different parallel channels, enabling efficient feature extraction. This mechanism is essential for optimizing decoding algorithms and enhancing the performance of BCI systems.

\subsection{BCI Signal Decoding}  
BCIs are primarily categorized into non-invasive systems, which place electrodes on the scalp but suffer from low signal-to-noise ratios due to transmission losses \cite{johnson2006signal}, and invasive systems, which record directly from the inner brain and have demonstrated remarkable effectiveness in decoding motor signals in animals \cite{velliste2008cortical}. The integration of CNNs and RNNs has significantly improved the accuracy of decoding invasive signals \cite{xie2018decoding,sliwowski2022decoding}. Emerging methods based on Transformers, such as swin-transformer \cite{chen2024subject} and other transformer alternatives \cite{he2021transformer}, have shown great potential in managing complex temporal dynamics in large datasets. However, increasing model complexity to handle high-sampling and multi-channel invasive BCIs usually results in substantial computational demands and power consumption, which poses challenges for clinical deployment on chips. SNNs, with their energy-efficient and biologically inspired operations, offer a promising solution. This study presents an SNN-based model for invasive BCI decoding that strikes a balance between low power consumption and manageable complexity.

\begin{figure*}[h!]
    \centering
    \includegraphics[width=0.95\linewidth]{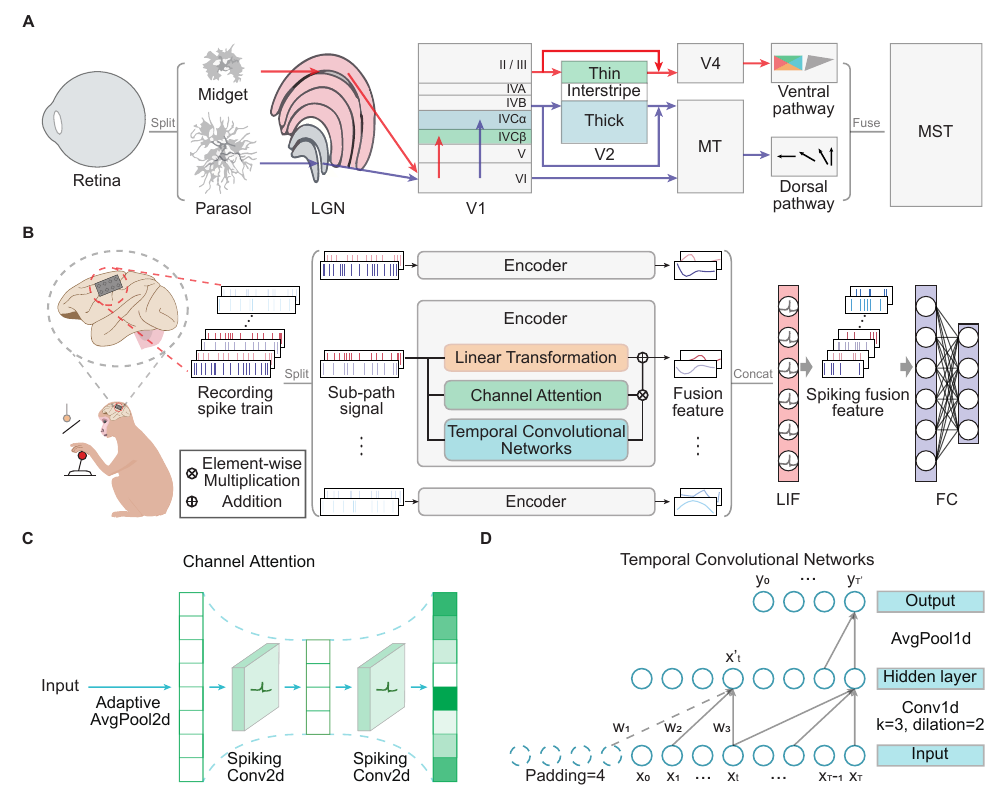}
    \caption{A. The diagram depicts the human brain's visual perception pathway, highlighting its parallel processing mechanisms and multiscale feature fusion. B. The overall MFSNN architecture, where the input consists of multi-channel neural signals from an invasive BCI in monkeys, and the final output represents monkey behavior categories. C. The channel attention module, which includes a bottleneck structure composed of spiking convolutional layers. D. The temporal convolutional network, which incorporates causal and dilated convolutions.}
    \label{fig1}
\end{figure*}

\section{Methods}

In this section, we provide a detailed description of our proposed method, including the model structure and its operational workflow.

\subsection{Overall Architecture}

Fig.\ref{fig1} B presents the overall architecture of our MFSNN algorithm. The input, a recorded spike train with \( N_{c} \) channels from an invasive BCI, is divided into \( N_{s} \) sub-path signals for parallel processing. Each sub-path is then processed by its corresponding sub-encoder for specialized analysis. Each sub-encoder comprises three key components: a linear transformation module for initial signal modification, a channel attention module for enhancing salient features across channels, and a temporal convolution network for extracting temporal dynamics. The outputs from these modules are designated as \( LT_{out} \), \( CA_{out} \), and \( TCN_{out} \), respectively. These outputs are integrated to produce the fusion feature \( E_i \) for the \( i \)th sub-encoder. The outputs from all sub-encoders, denoted as \( E_1 \) through \( E_{N_{s}} \), are then concatenated to form \( E_{out} \). This consolidated feature vector is passed through a spiking classifier, transforming it into a spiking fusion feature and decoding it to yield the classification result \( Net_{out} \).

In the subsequent sections of this paper, we will provide a comprehensive explanation of the functional roles and contributions of each module within the MFSNN framework, highlighting their synergistic impact on the network's overall performance in addressing cross-day decoding challenges with high-fidelity neural signal decoding for invasive BCI systems.

\subsection{Sub-Encoder}

\subsubsection{Linear Transformation (LT)}

The given input neural signal is denoted as \( {Input} \in \mathbb{R}^{C \times 1 \times T} \), where \( C \) represents the channel dimension, jointly determined by the number of raw data channels \( N_{c} \) and the number of sub-encoders \( N_{s} \), shown as following equations:

\begin{equation}
C = N_{c}/N_{s}
\text{.}
\label{eq:C}
\end{equation}

At the outset of our processing procedure, a learnable matrix \(M_{l}\) for linear transformation (LT) is implemented. This LT reduces the sequence length from \( T \) to \( T^{'} \), generating the output \( {LT_{out}} \in \mathbb{R}^{C \times 1 \times T'} \). 

\begin{equation}
LT_{out} = Input * M_{l}
\text{.}
\label{eq:LT}
\end{equation}

This critical step here is to capture the representative features at the raw data level of the sequence, thereby establishing a foundation for the subsequent multiscale feature fusion within the network's architecture.

\subsubsection{Channel Attention (CA)}

In invasive BCI neural signals, each channel captures the activity of different neuronal populations, which contribute variably to the same task. The sub-encoder employs a spiking Channel Attention (CA) module to extract spatial features from the sub-path signal. As depicted in Fig.\ref{fig1} C, the CA module first provides adaptive global average pooling \( F_{ap}(\cdot) \) to the \( {Input} \in \mathbb{R}^{C \times 1 \times T} \) to capture spatial features \( f_{s}\in \mathbb{R}^{C \times 1 \times 1} \) across channels.

\begin{equation}
f_{s}[C,1,1] = F_{ap}(Input) = \frac{1}{T} \sum_{i=1}^{T} Input[C,1,T] \
\text{.}
\label{eq:fs}
\end{equation}

Subsequently, a bottleneck structure, composed of two spiking convolutional layers \( SConv2d(\cdot) \) and modified with Leaky Integrate-and-Fire (LIF) neuron-based activation functions \( LIF(\cdot) \), is incorporated to compress features first and then expand to derive the channel weight distribution \( \tilde{w}_{c} \). 

\begin{equation}
\begin{aligned} 
SConv2d(\cdot) = LIF(Conv2d(\cdot))
\\ \tilde{w}_{c} = SConv2d(SConv2d(f_{s}))
\text{.}
\label{eq:CA}
\end{aligned}
\end{equation}

This mechanism allows the model to focus on some specific channels those are more critical for decoding target tasks while effectively suppressing redundancy and noise in the signals. Consequently, this approach enables more precise channel selection, thereby enhancing the accuracy of predictions.

\subsubsection{Temporal Convolution Network(TCN)}

Invasive BCI signals are temporal sequences, making the features along the time dimension particularly important. The sub-encoder utilizes a Temporal Convolution Network (TCN) to effectively capture long-term dependencies along with the time dimension through causal convolution and dilated convolution. As shown in Fig.\ref{fig1} D, taking a single channel from the \( {Input} \in \mathbb{R}^{C \times 1 \times T} \) for example, the temporal sequence \( X = (x_1, x_2, ..., x_T) \) is subjected to a time convolution with a kernel size of \( k = 3 \), a dilation rate of \( d = 2 \), and padding defined as \( (k - 1) \times d + 1 = 4 \). The output dimension of the hidden layer remains unchanged. For any moment \( x'_t \) in the hidden layer output \( X' = (x'_1, x'_2, ..., x'_T) \) and its corresponding convolution kernel \( F_t = (w_1, w_2, w_3) \), the following equation holds:
\[ x'_t = b + \sum_{i=1}^{3} w_i \times x_{t - (3-i) \times d} \]

Then, an average pooling operation with a window size of \( p \), denoted as \( F_{ap}(\cdot) \), is applied to \( X' \) to obtain the module output \( Y = (y_0, ..., y_{T'}) \) for the temporal sequence of a single channel. 

\begin{equation}
\begin{aligned} 
Y(y_0, ..., y_{T'}) &= F_{ap}(X'(x'_1, x'_2, ..., x'_T))\\
y_t &= \frac{1}{p} \sum_{i=t-p+1}^{t} x'_t, \ \ p = \frac{T}{T'}
\text{.}
\label{eq:TCN}
\end{aligned}
\end{equation}

By performing the aforementioned temporal convolution simultaneously across the \( C \) channels of the sub-path signal \( {Input} \in \mathbb{R}^{C \times 1 \times T} \), the module output is obtained as \( {TCN_{out}} \in \mathbb{R}^{C \times 1 \times T'} \).

Compared to traditional models for processing temporal signals such as RNN, GRU, and Transformer, the proposed TCN offer superior parallel processing capabilities. All convolutional operations can be computed simultaneously, thereby enhancing computational efficiency and processing speed.

\subsubsection{Feature fusion}

We integrate the raw data-level features \( {LT_{out}} \in \mathbb{R}^{C \times 1 \times T^{'}} \) generated by the LT module, the spatial features \( \tilde{w}_{c} \in \mathbb{R}^{C \times 1 \times 1} \) produced by the CA module, and the temporal features \( {TCN_{out}} \in \mathbb{R}^{C \times 1 \times T'} \) derived from the TCN module to obtain the integrated features of \( i \)th sub-encoder \( E_{i} \in \mathbb{R}^{C \times 1 \times T'} \). 

\begin{equation}
E_i = LT_{out} + \tilde{w}_{c} * TCN_{out}
\text{.}
\label{eq:Encoder}
\end{equation}

Subsequently, we concatenate the outputs of all \( N_s \) sub-encoders to form the output of the entire signal encoder. The resulting output is represented as \( E_{out} \in \mathbb{R}^{N_c \times 1 \times T'} \).

\begin{equation}
E_{out} = Concat(E_i) \qquad (i=1,2,3,...,N_s)
\text{.}
\label{eq:Eout}
\end{equation}

\subsection{Spiking Classifier}

The spiking classifier is composed of a LIF neuron layer and a fully connected layer, which takes the fused feature \(E_{out}\) as the input current, leading to fluctuations in membrane potential and the generation of spikes. Utilizing a spiking layer to extract sparse spike features, the classifier then decodes to achieve the classification result \(Net_{out}\). Moreover, the spiking classifier can generalize with fine-tuning on small samples after pre-training. Its advantages include reduced computational costs and enhanced model adaptability, making it particularly suitable for addressing cross-day decoding issues and for future deployment on neuromorphic chips.

\begin{figure*}[h!]
    \centering
    \includegraphics[width=1\linewidth]{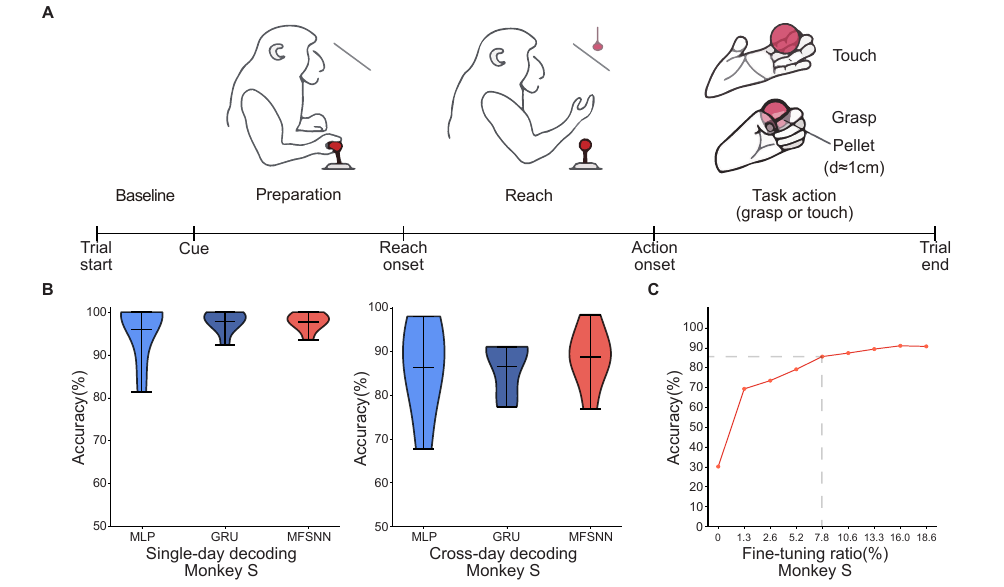}
    \caption{A. The single-hand grasp-and-touch experimental paradigm. B. MLP, GRU, and MFSNN single-day and cross-day decoding experiments on neural signals of monkey S. C. The accuracy of different fine-tuning ratios.}
    \label{fig2}
\end{figure*}

\section{Experiments and Results} 

\subsection{Experimental Setting}

\subsubsection{Data Details}

Dataset 1: The experimental paradigm is depicted in Fig.\ref{fig2} A. The macaque monkeys are used as subjects, with each trial divided into four phases. Baseline—the task initiation is marked by the monkey pulling the joystick for 1 second until the appearance of a cue. Preparation—following the cue, the target object appears between 0.1 to 0.5 seconds, after which the monkey releases the joystick. Reach—the monkey releases the joystick and reaches for the position of the ball within approximately 0.5 to 1 second. Task action—the task concludes upon completion of the touch or grasp action, which lasts over 1 second. Each trial lasts 2 to 4 seconds in total. The tasks are categorized into four types: right-hand touch, right-hand grasp, left-hand touch, and left-hand grasp, with each trial conducted independently. Concurrently, 128-channel neural signals from the monkey's motor cortex (M1) are recorded during the trial at a sampling rate of 30 kHz. The experimental data were collected over eight separate days, from 01/26/2022 to 03/09/2022, with an average of approximately 300 trials per day.

Dataset 2: The experimental paradigm is depicted in Fig.\ref{fig3} A. The dataset from the research conducted by Churchland et al.\cite{churchland2012neural}, pertains to the collection of neural signals extracted from two rhesus monkeys, identified as J and N, during their performance of the center-and-out task to eight directions. The signals were recorded using a pair of 96-channel electrode arrays implanted in the M1, with an average of 2,000 trials per day.

\subsubsection{Implementation Details}

All training and testing were conducted on two NVIDIA GeForce RTX 4060 graphics cards. Within the MFSNN network, the spiking neuron model employs the LIF neuron model, with a time window set to 20ms. During the training process, we train the MFSNN with the Adam optimizer and a batch size of 32. The learning rate was dynamically adjusted using the cosine annealing learning rate schedule, starting from 0.01 and ranging down to 0.0001. In comparative experiments, we adopted a similar training strategy for MLP and GRU.

\subsection{Neural Signal Decoding}
\subsubsection{Single-hand Grasp-and-touch Task}

In the study of neural signal decoding for single-hand grasp-and-touch task, we focus our analysis on decoding signals from the ``Task action" phase. Acknowledging the variability of neural signal characteristics over time, we employ two testing methods: single-day and cross-day decoding. The single-day decoding uses training and testing sets from the same day with an 8:2 ratio; the cross-day decoding uses sets from different days. The data includes eight days spanning from January 26 to March 9, 2022. The single-day decoding experiments were conducted daily, amounting to a total of eight sets.

The cross-day decoding experiment is further divided into two parts: one part trained with data from January 26 and tested with data from January 30 to February 9. The left data were trained with data from March 3 and tested with data from March 6 to 9. Each part consisted of three experiments, making a total of six cross-day decoding tests. During the cross-day decoding, a small proportion of the test set is fine-tuned to enhance the generalization ability of all models.

We compare the performance of three algorithms(MLP, GRU, and MFSNN) under the two testing paradigms. As shown in Fig.\ref{fig2} B, the average accuracy of the three algorithms under single-day decoding is comparable and notably high($>$95\%), attributed to the stability of neural signal feature distribution within the same day. That is also why the accuracy rates of single-day decoding are higher than those of cross-day decoding. In cross-day decoding, despite the same fine-tuning apply to MLP and GRU as MFSNN, MFSNN still show a higher accuracy rate($>$80\%), demonstrating its superior decoding capability in the face of cross-day changes in neural signal characteristics.

Furthermore, we conduct gradient testing on the ratio of fine-tuning data for MFSNN in cross-day decoding to analyze its impact on performance. As shown in Figure 2.C, when the fine-tuning data ratio reaches 7.8\%, the model performance could be stabilized and good enough($>$80\%). This slightly higher fine-tuning ratio may be due to the minimal differences among the four types of actions in the task, resulting in more similar neural signals, thus requiring more fine-tuning data to achieve good generalization.

\subsubsection{Center-and-out Task}

In the center-and-out task experiments conducted on monkeys J and N, we test both single-day and cross-day decoding. The single-day decoding task is similar to the single-hand grasp-and-touch task. For cross-day decoding, the model is trained on the first day's data and test on the remaining days. For example, monkey J had data from four days. The network will be trained on the first day and tested on the second, third, and fourth days. 

\begin{figure*}[htbp!]
    \centering
    \includegraphics[width=0.93\linewidth]{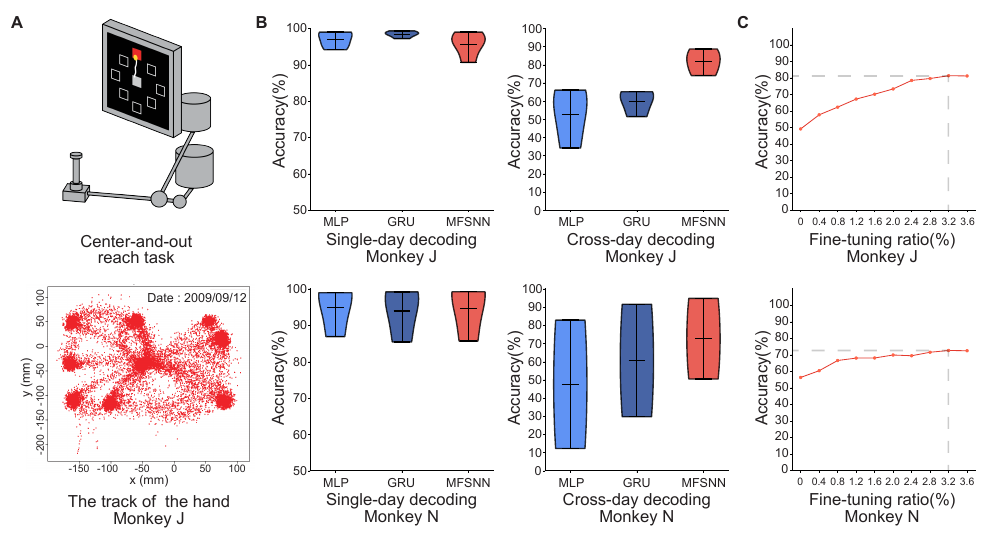}
    \caption{A. Center-and-out experimental paradigm. B. MLP, GRU, and MFSNN single-day and cross-day decoding experiments on neural signals of monkey J and N. C. The accuracy of different fine-tuning ratios.}
    \label{fig3}
\end{figure*}

The results, shown in Fig.\ref{fig3} B, indicate that all three models achieved very high accuracy rates ($\ge$95\%) in single-day decoding, with no significant differences among them. In cross-day decoding, under the same fine-tuning conditions, average accuracy rate of MFSNN is significantly higher than that of MLP and GRU, and the results are more concentrated, indicating stronger generalization and higher robustness.

We also conduct a gradient test of the fine-tuning data ratio for MFSNN on both monkeys to assess its impact on performance. As shown in Fig.\ref{fig3} C, with a fine-tuning data ratio of only 3.2\%, stable and effective generalization is achieved in both monkeys.

In summary, comparing the results of the two experiments, we find that due to the short time span of single-day data, the distribution of neural signal features do not change significantly, resulting in high and similar accuracy rates for all models. However, in cross-day decoding, the significant changes in the distribution of neural signal features due to the longer time span led to a decrease in the average accuracy rate of all models. Nevertheless, MFSNN, with its excellent multi-level feature fusion mechanism, can stably capture the distribution of neural signals on different days, and with a small sample ($<$8\%) of fine-tuning, it can achieve efficient and robust decoding performance.

\begin{figure}[htbp!]
    \centering
    \includegraphics[width=0.9\linewidth]{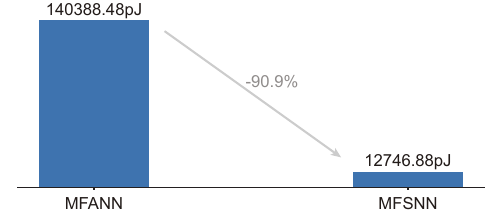}
    \caption{Energy consumption for processing a single data entry.}
    \label{fig4}
\end{figure}

\subsection{Energy Consumption}

We estimate the theoretical energy consumption of MFSNN and Multiscale Fusion enhanced Artificial Neural Network(MFANN) by the following two equations\cite{mark2014computing,yao2024spike}:

\begin{equation*}
SOP_{s}(l) = Rate \times T \times FLOP_{s}(l)
\end{equation*}
\begin{equation}
E_{MFSNN} = E_{AC} \times \sum_{i=1}^{16}(SOP_{LT}^i + SOP_{CA}^i + SOP_{TCN}^i)
\text{.}
\label{eq:E}
\end{equation}

\(SOPs(l)\) means synaptic operations (the number of spike-based accumulate(AC) operations) of layer \(l\), Rate is the average firing rate of input spike train to layer \(l\), \(T\) is the time window of LIF neurons, and \(FLOPs(l)\) refers to the floating point operations (the number of multiply-and-accumulate (MAC) operations) of layer \(l\). We assume that the MAC and AC operations are implemented on the 45nm hardware\cite{mark2014computing}, with \(E_{MAC} = 4.6pJ\) and \(E_{AC} = 0.9pJ\).

\begin{figure}[htbp!]
    \centering
    \includegraphics[width=0.9\linewidth]{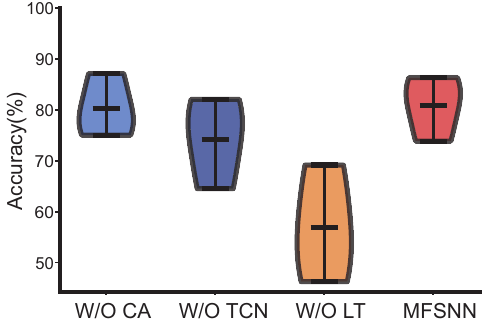}
    \caption{The ablation experiments of MFSNN.}
    \label{fig5}
\end{figure}

Under the cross-day decoding test paradigm of dataset 2 with monkey J, the computational energy consumption on 45nm hardware for a single spike train is simulated for MFANN and MFSNN. The results, as shown in Fig.\ref{fig5}, indicate that the energy consumption of MFSNN is reduced by 90.9\% compared to that of the similarly structured MFANN.

\subsection{Ablation Study}

In the ablation study conduct on the cross-day decoding experiment of monkey J from dataset 2, we scrutinize the roles of the CA, TCN, and LT modules within the MFSNN. Analysis of Fig.\ref{fig5} reveals that all three modules significantly contribute to the model's performance. Particularly, the contributions of LT and TCN are substantial. Given the temporal nature of neural signals, the feature extraction along the temporal dimension by TCN is of paramount importance. LT, on the other hand, focuses on extracting deep features from the raw data, capturing nuances that may be overlooked by temporal and spatial features. These two modules serve as two critical tiers in the multi-level feature fusion strategy of the MFSNN, essential for the model's overall performance.

Upon examining Fig.\ref{fig5}, it is observed that the performance distribution of MFSNN without the CA module is represented by a violin plot that is wider at the bottom and narrower at the top, whereas the inclusion of the CA module reverses this, presenting a plot that is narrower at the bottom and wider at the top. This indicates that although the CA module has a limited effect on enhancing the average accuracy, it effectively elevates a number of cross-day decoding outcomes from below to over 80\%, a change that holds significant practical implications.

In summary, the results of the ablation study demonstrate that the CA, TCN, and LT modules, which target different aspects of feature extraction, collectively underpin the efficacy of the MFSNN.

\section{Conclusion} 

Stable and long-term decoding of neural signals is crucial for BCIs, which is an advanced fusion of neuroscience and artificial intelligence.  In this paper, we propose the Multiscale Fusion enhanced Spiking Neural Network (MFSNN) framework, which emulates the parallel processing and multiscale feature fusion in human visual perception, enabling real-time and energy-efficient neural signal decoding. The proposed SNN-based model decodes high-throughput invasive brain signals with reduced energy consumption, offering a practical solution for invasive BCI systems. Our experiments demonstrate that it is feasible and robust to MFSNN for cross-day decoding in the BCI system. 

From Fig.\ref{fig2} C and Fig.\ref{fig3} C, it is evident that the cross-day decoding based on the MFSNN performance is affected by the ratio of fine-tuning trials. The decoding accuracy of the model initially improves as the fine-tuning ratio increases and then stabilizes. This indicates that the model, after generalizing over a certain amount of data, can easily achieve good decoding performance. However, the required proportion of fine-tuning data is influenced by the complexity of the dataset and the difficulty of the task paradigm. According to the hypothesis of neural manifold \cite{gallego2020long,zhao2024enhanced}, there is frequently a latent low-dimensional stable neural manifold within high-dimensional neural signal data. In our next steps, we will adopt this perspective, aiming to design a long-term stable decoding algorithm that first reduces the dimensionality of microscopic spike signals to the neural manifold and then decodes macroscopic behavioral signals from the neural manifold.

The MFSNN’s multiscale feature fusion framework is also well-suited for implementing neuromorphic chips, offering an energy-efficient solution for the online decoding of invasive BCI signals. Regarding the deployment of algorithms on hardware, this paper explores the feasibility of energy consumption estimation on neuromorphic chips.

\bibliography{aaai25.bib}

\end{document}